\documentclass[runningheads]{llncs}
\usepackage{multirow}
\usepackage{makecell}
\usepackage{graphicx}
\usepackage{enumitem}

\begin{document}

\title{Deep neural networks approach to microbial colony detection---a comparative analysis}
\titlerunning{DNN approach to microbial colony detection---a comparative analysis}
%
\author{Sylwia Majchrowska\inst{1,2} \and
Jarosław Pawłowski\inst{1,2} \and
Natalia Czerep\inst{1,3} \and
\\Aleksander Górecki\inst{1,2} \and
Jakub Kuciński\inst{1,3} \and
Tomasz Golan\inst{1}}
\authorrunning{S. Majchrowska et al.}
%
\institute{NeuroSYS, Rybacka 7, 53-656 Wrocław, Poland,
\and
Wroclaw University of Science and Technology,\\
Wybrzeże S. Wyspiańskiego 27, 50-372 Wroclaw, Poland \and
University of Wroclaw, Fryderyka Joliot-Curie 15, 50-383 Wroclaw, Poland}

\maketitle              
\begin{abstract}
Counting microbial colonies is a fundamental task in microbiology and has many applications in numerous industry branches.
Despite this, current studies towards automatic microbial counting using artificial intelligence are hardly comparable due to the lack of unified methodology and the availability of large datasets.
The recently introduced AGAR dataset is the answer to the second need, but the research carried out is still not exhaustive.
To tackle this problem, we compared the performance of three well-known deep learning approaches for object detection on the AGAR dataset, namely two-stage, one-stage and transformer based neural networks. The achieved results may serve as a benchmark for future experiments.

\end{abstract}

\section{Introduction}
\label{sec:introduction}

The ability to automatically and accurately detect, localize, and classify bacterial and fungal colonies grown on solid agar is of wide interest in microbiology, biochemistry, food industry, or medicine. An accurate and fast procedure for determining the number and the type of microbial colonies grown on a Petri dish is crucial for economic reasons – industrial testing often relies on proper determination of colony forming units (CFUs). Conventionally, the analysis of samples is performed by trained professionals, even though it is a time-consuming and error-prone process. To avoid these issues, automated methodology based on artificial intelligence can be applied.

A common way of counting and classifying objects using deep learning (DL) is to first detect them and then count the found instances distinguishing between different classes. We compared the results of microbe colonies counting using selected detectors belonging to different classes of neural network architectures, namely two-stage, one-stage, and transformer based models

Our experiments were conducted on the Annotated Germs for Automated Recognition (AGAR) dataset with \textit{higher-resolution} subset~\cite{bib:AGAR}, which consists of around 7k annotated Petri dish photos with five microbial classes. This paper focuses on setting benchmarks for the detection and counting tasks using state-of-the-art (SoTA) models.

\section{Object detection}
\label{sec:object_review}

Object detection was approached using two major types of DL-based architectures, namely two-stage and one-stage models. \textbf{Two-stage} detectors find class-agnostic object proposals in the first stage, and in the second stage the proposed regions are assigned to the most likely class. They are characterised by high localization and classification accuracy. The largest group of two-stage models are Region Based Convolutional Neural Networks (R-CNN)~\cite{girshick2014rich}, whose main idea is based on extracting region proposals from the image. Over the years, the networks from this family have undergone many modifications. In the case of \textit{Faster R-CNN}~\cite{bib:Faster2015}architecture, a Region Proposal Network (RPN) was used instead of the Selective Search algorithm. This allows for significant reduction of the model's inference time. In order to reduce issues with over-fitting during training, \textit{Cascade R-CNN}~\cite{bib:Cascade2018} was introduced as multi-stage object detector, which consists of multiple connected detectors that are trained with increased intersection over union (IoU) thresholds. A year later, authors of \textit{Libra R-CNN}~\cite{pang2019libra}  focused on balancing the training process by IoU-balanced sampling, balanced feature pyramid, and balanced L1 loss.
In the following years, researchers used Faster R-CNN while replacing its backbone (CNN used as feature extractor) with newer architectures. The most recent concept is \textit{Composite Backbone Network V2}~\cite{liang2021cbnetv2} (CBNetV2), which groups multiple pre-trained backbones of the same kind for more efficient training.

On the other hand, the \textbf{one-stage} architectures are designed to directly predict classes and bounding box locations. For this reason, one-stage detectors are faster, but usually have relatively worse performance. Single-stage detection was popularized in DL mainly by You Only Look Once models (YOLO v1~\cite{redmon2016you}, 9000~\cite{redmon2017yolo9000}, v3~\cite{redmon2018yolov3}, v4~\cite{bochkovskiy2020yolov4}), primarily developed by Joseph Redmon. Recently, the authors of \textit{YOLOv4}~\cite{bochkovskiy2020yolov4} have enhanced the performance of YOLOv3 architecture using methods such as data augmentation, self-adversarial training, and class label smoothing, all of which improve detection results without degrading the inference speed. Moreover, the authors of \textit{EfficientDet}~\cite{efficientdet2020} introduce changes which contribute to an increase in both accuracy and time-performance of object detection. The main proposed changes include using weighted Bidirectional Feature Pyramid Network (BiFPN), compound scaling, and replacing the backbone network with EfficientNet~\cite{efficientnet2019}.
EfficientNet as a backbone connects with the idea of scalability from compound scaling, which allows the model to be scaled to different sizes and to create a family of object detectors for specific uses.

Additionally, the \textbf{transformers} have recently become a next generation of neural networks for all computer vision applications, including object detection~\cite{bib:Zhu2020,khan2021transformers,elnouby2021xcit}. The interest in replacing CNN with transformers is mainly due to their efficient memory usage and excellent scalability to very large capacity networks and huge datasets. The parallelization of transformer processes is achieved by using an attention mechanism applied to images split into patches treated as tokens. The utilization of the transformer architecture to generate predictions of objects and their position in an image was first proposed in DEtection TRansformer (DETR) network ~\cite{carion2020endtoend}. The architecture uses a CNN backbone to learn feature maps, then feeds transformer layers. In comparison to DETR, \textit{Deformable DETR}~\cite{bib:Zhu2020} network replaces self-attention in the encoder and cross-attention in the decoder with multi-scale deformable attention and cross-attention. Deformable attention modules only attend to a small set of key sampling points around a reference point which highly speeds up the training process. The recently introduced \textit{Cross-Covariance Image Transformers} (XCiT)~\cite{elnouby2021xcit} concept is a new family of transformer models for image processing. The idea is to use a transformer based neural network as a backbone for two-stage object detection networks. XCiT splits images into fixed size patches and reduces them into tokens with a greater number of features with the use of a few convolutional layers with Gaussian Error Linear Units (GELU)~\cite{hendrycks2020gaussian} in between. The idea behind the model is to replace self-attention with transposed attention (which is over feature maps instead of tokens).

\section{AGAR dataset}

The AGAR dataset~\cite{bib:AGAR} contains images of microbial colonies on Petri dishes taken in two different environments, which produced \textit{higher resolution} and \textit{lower resolution} images. The differences are between the lighting conditions and apparatuses. \textit{Higher resolution} images, which were used in our studies, can be divided into \textit{bright}, \textit{dark} and \textit{vague} subgroups. On the other hand, considering the number of colonies, samples can be defined as \textit{empty}, \textit{countable} and \textit{uncountable}. The dataset includes five classes, namely \textit{E.coli, C.albicans, P.aeruginosa, S.aureus, B.subtilis}, while annotations are stored in json format with the information about the number and type of microbe, environment and coordinates of bounding boxes.

In this paper, we present results of experiments performed using a subset of the AGAR dataset, which consists of 6990 images in total. In our case only \textit{higher resolution} (mainly $4000\times4000$~px), \textit{dark} and \textit{bright}, without \textit{vague}, samples with \textit{countable} number of colonies were chosen. Firstly images were split into train and validation subsets (the same for each experiment), and then divided into $512\times512$~px patches as described in~\cite{bib:AGAR}. At the end---in the test stage---whole images from validation subset of the Petri dish were used (for detailed description of the procedure see Supplementary materials from ~\cite{bib:AGAR}).

\section{Benchmarking methodology}
\label{sec:benchmarks}

We compared the performance of selected models using several metrics: architecture type and size, inference time, and detection and counting accuracy.

During time measurements, the inference was executed on GeForce GTX 1080 Ti GPU using the same patch with 6 ground truth instances. The models were first loaded into memory, then inferred 100 times sequentially (ignoring the first 20 times for warming up) to calculate averaged time and its standard deviation for each model separately.

As to detection results, the detector performance was evaluated twofold – by measuring the effectiveness of detection and counting. As an evaluation metric for colony detection, we rely on the mean Average Precision (mAP), to be precise mAP@.5:.95, averaged over all 5 classes. The efficiency of colony counting was measured based on Mean Absolute Error (MAE), and Symmetric Mean Absolute Percentage Error (sMAPE).

With the growing popularity of DL, many open source software libraries implementing SoTA object detection algorithms emerge. Results provided for Faster R-CNN and Cascade R-CNN were taken from~\cite{bib:AGAR} for comparison purposes. Similarly, in our experiments we relied on MMDetection~\cite{mmdetection} framework (Libra R-CNN, CBNetV2, Deformable DETR, XCiT), Alexey Bochkovskiy’s Darknet-based implementation of YOLOv4~\cite{bochkovskiy2020yolov4}, and Ross Wightman’s PyTorch~\cite{effdet-pytorch} reimplementation of official EfficientDet’s TensorFlow implementation. To perform model training, we used the default parameters as for COCO dataset in the above mentioned implementations. In case of YOLOv4, we changed the input size to $512\times512$~px in order to match the size of the generated patches. We used pretrained backbones in all experiments. Traditional two- and one-stage networks were trained with Stochastic Gradient Descent (SGD) optimizer, as opposed to Transformer based architectures, where AdamW~\cite{loshchilov2019decoupled} optimizer was used. The values of initial learning rate vary between $10^{-3}$ and $10^{-5}$ for each model. All networks were trained until loss values saturated for validation subset. We also chose commonly used augmentation strategies of selected models, like flips, crops and resizes of images.

\subsection{Results}

\begin{figure}[!hbt]
\centering
\includegraphics[width=.9\textwidth]{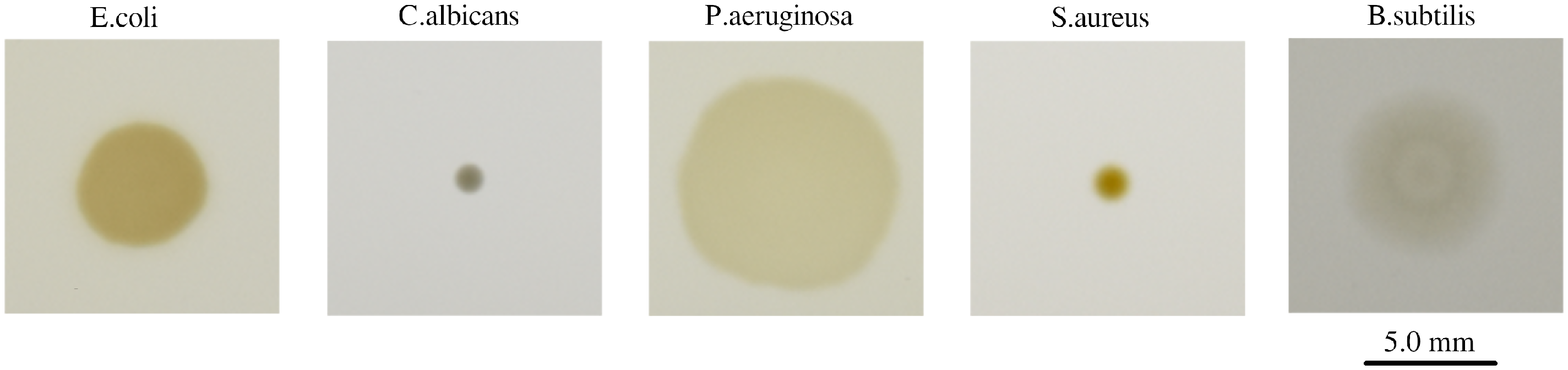}
\includegraphics[width=.9\textwidth]{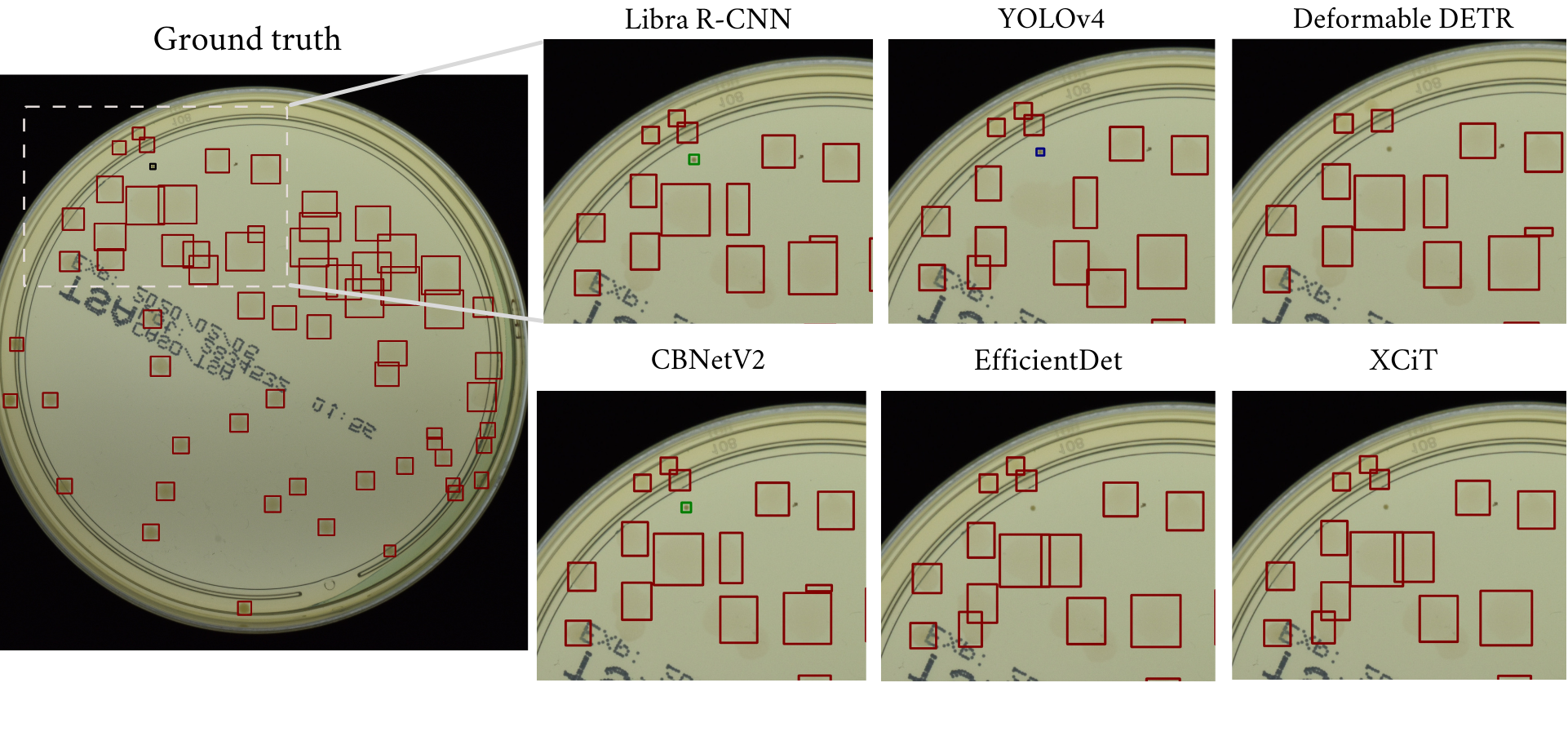}

\caption{\label{fig:examples}Examples of images of microbial species and achieved predictions for a selected sample. Whole image of the Petri dish presents ground truth annotations, while the white dashed rectangle indicates the region chosen for visualisation of predicted results. Red rectangles mark \textit{P.aeruginosa} species, the black one -- \textit{Contamination} (microorganism not intentionally cultured), green -- \textit{C.albicans}, navy -- \textit{S.aureus}.}
\end{figure}

Mean averaged precisions presented in Table~\ref{tab:mAP} are averaged over the all microbe classes. Calculated value of mAP@.5:.95 varies between 0.491 and 0.529. The most efficient results in terms of accuracy and inference speed were achieved for YOLOv4 architecture. On the other hand, transformer based architectures present slightly worse performance. Some interesting cases were presented in Fig.~\ref{fig:examples}. The selected image presents the same microbial species (\textit{P.aeruginosa}), which forms two different sizes of colonies due to agar inhomogeneities, making detection even more challenging. Labeled small contamination is not perceived by all models (transformer based and EfficientDet-D2), and some of them (YOLOv4, Deformable DETR) also have problems with precise localization of blurred colonies. Two-stage detectors have a tendency to produce some excessive predictions.

{
\begin{table}
\centering
\caption{Benchmarks for tested models on the \textit{higher-resolution} subset of AGAR dataset. The model size is given in terms of number of parameters (in millions). In case of XCiT model number of backbone's parameters is given in brackets.
}
\label{tab:mAP}
\begin{tabular}{p{2cm}p{2.8cm}p{2.2cm}p{1.1cm}p{1.8cm}>{\centering\arraybackslash}p{1.4cm}}
\hline
\textbf{Type}   & \textbf{Model}   & \textbf{Backbone}  &\textbf{mAP .5:.95} & \textbf{Inference time~(ms)} & \textbf{Size (M)}
\\ \hline
\multirow{4}{*}{two-stage}   & Faster R-CNN     &   ResNet-50        &   0.493            &    $54.06 \pm 1.72$   & 42 \\
   & Cascade R-CNN    &   ResNet-50        &   0.516            &    $76.31 \pm 1.96$   & 69   \\
   & Libra R-CNN      &   ResNet-50        &   0.499            &    $33.34 \pm 0.49$   & 41     \\
   & \multirow{2}{*}{CBNetv2}          &   Double &   \multirow{2}{*}{0.502}            &    \multirow{2}{*}{$43.79 \pm 0.42$}   & \multirow{2}{*}{69} \\
 &   &  ResNet-50 & &   \\ \hline
\multirow{2}{*}{one-stage}   & YOLOv4           &   CSPDarknet53     &   0.529            &    $17.46 \pm 0.17$  & 64    \\
   & EfficientDet-D2  &   EfficientNet-B2  &   0.512            &        $45.59 \pm 1.06$    &      8        \\ \hline
transformer & Deformable DETR  &   ResNet-50        &   0.492            &    $72.40 \pm 0.65$   & 40      \\
transformer & \multirow{2}{*}{Faster R-CNN}  &   \multirow{2}{*}{XCiT-T12} &   \multirow{2}{*}{0.491}            &    \multirow{2}{*}{$110.99 \pm 3.85$}  & 25   \\ 
backbone & & & & & (7)\\ \hline
\end{tabular}
\end{table}
}

\begin{figure}[!ht]
\centering
\includegraphics[width=\textwidth]{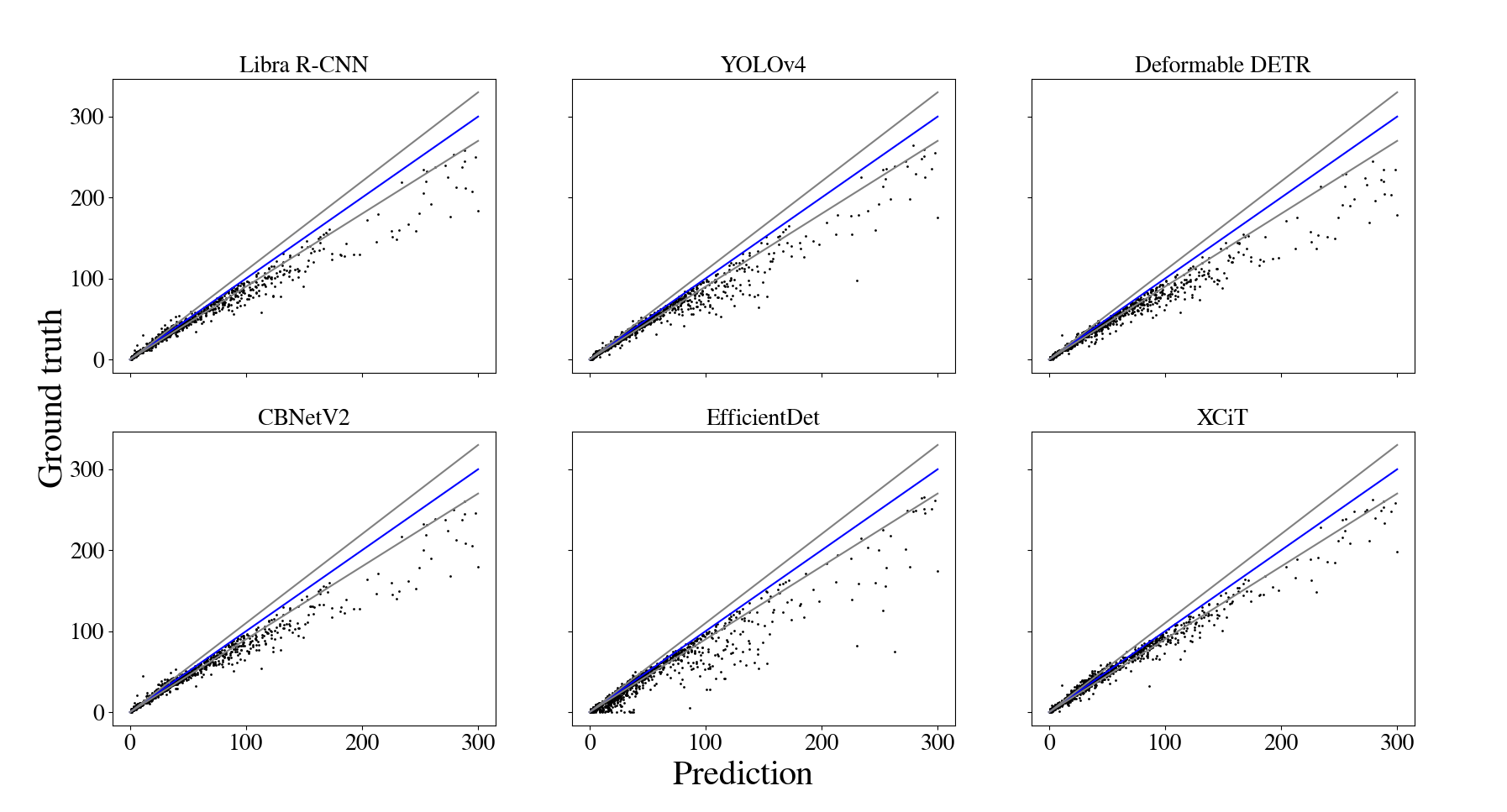}
\caption{\label{fig:gt_vs_pred_types}The performance of microbial colony counting for 6 different models.}
\end{figure}

\pagebreak
The performance of selected architectures for microbial counting is presented in both Table~\ref{tab:counting} and Fig.~\ref{fig:gt_vs_pred_types}, while Table~\ref{tab:bacteria} shows all five microbial species separately. In general, all detectors perform better for microbes that form clearly visible, separate colonies. The biggest problem with locating individual colonies was observed for \textit{P.aeruginosa}, , where the tendency for aggregation and overlapping is the greatest. Overall, the best results were obtained for the YOLOv4 model, where the predicted count of microbial colonies is the closest to ground truth in range from 1 to 50 instances (see Fig.~\ref{fig:gt_vs_pred_types}) -- the most operable scope for industrial applications. The worst performance was observed for the EfficientDet-D2 model -- where small instances of microbial colonies were omitted (not localized at all), which may be caused by resizing patches to fit the input layer size. Very low contrast between the agar substrate and the colony (\textit{bright} subset of AGAR dataset) is an additional problem here.

{
\begin{table}
\centering
\caption{Symmetric Mean Absolute Percentage Error (sMAPE) and Mean Absolute Error (MAE) obtained for different models.}
\label{tab:counting}
\begin{tabular}{p{3cm}p{3cm}p{1.2cm}p{1.2cm}}
\hline
\textbf{Model}   & \textbf{Backbone}  & \textbf{MAE} & \textbf{sMAPE}\\\hline
Faster R-CNN     &   ResNet-50        &     4.75     &  5.32\% \\
Cascade R-CNN    &   ResNet-50        &     4.67     &  5.15\% \\
Libra R-CNN      &   ResNet-50        &     4.21     &  5.19\% \\
CBNetv2          &   Double ResNet-50 &     4.49     &  5.23\% \\ \hline
YOLOv4           &   CSPDarknet53     &     4.18     &  5.17\% \\
EfficientDet-D2  &   EfficientNet-B2  &     5.66     &  10.81\% \\ \hline
Deformable DETR  &   ResNet-50        &     4.82     &  5.30\% \\
Faster R-CNN     &   XCiT-T12         &     3.53     &  6.30\% \\ \hline
\end{tabular}
\end{table}
}

{
\begin{table}
\centering
\caption{Symmetric Mean Absolute Percentage Error (sMAPE) obtained for different types of microbes.}
\label{tab:bacteria}
\begin{tabular}{p{3cm}p{1.5cm}p{2cm}p{2.2cm}p{1.3cm}p{1.3cm}}
\hline
\textbf{Model}&\textit{E.coli}&\textit{C.albicans}&\textit{P.aeruginosa}&\textit{S.aureus}&\textit{B.subtilis} \\ \hline
Faster R-CNN &   \multirow{2}{*}{5.40\%} &  \multirow{2}{*}{2.98\%}  &   \multirow{2}{*}{5.53\%} &  \multirow{2}{*}{2.78\%}  &   \multirow{2}{*}{1.96\%} \\
ResNet-50 &   &   &  &  &  \\
Cascade R-CNN          &  4.40\% & 2.52\%  &  5.04\% & 2.51\%  &  1.36\% \\
Libra R-CNN            &  4.15\% & 8.49\%  & 14.57\% & 3.42\%  &  1.66\% \\
CBNetv2                &  6.06\% & 6.03\%  &  8.74\% & 2.80\%  &  2.57\%  \\ \hline
YOLOv4                 &  5.13\% & 1.99\%  &  6.43\% & 2.40\%  &  1.12\% \\
EfficientDet-D2        &  3.27\% &  4.53\% &  5.59\% & 3.49\%  &  1.21\%\\ \hline
Deformable DETR        &  2.95\% & 2.56\%  &  9.02\% & 2.33\%  &  1.44\% \\
Faster R-CNN     &  \multirow{2}{*}{5.33\%} &  \multirow{2}{*}{5.68\%} &  \multirow{2}{*}{9.12\%} &  \multirow{2}{*}{2.24\%} &  \multirow{2}{*}{1.87\%} \\
XCiT     &  &   &  & & \\ \hline
\end{tabular}
\end{table}
}

\newpage
\section{Conclusions}
\label{sec:conclusions}
In the conducted studies, we analyzed eight SoTA deep architectures in terms of model type, size, average inference time, and the accuracy of detecting and counting microbial colonies from images of Petri dishes. A detailed comparison was performed on AGAR dataset~\cite{bib:AGAR}.

The presented results do not differ much between the different types of architectures. It is worth noting that we chose rather smaller, typical backbones for the purposes of this comparison to create a baseline benchmark for different types of detectors. It appeared that the most accurate (\mbox{$\mathrm{mAP}=0.529$}) and the fastest model (17 ms) is one-stage YOLOv4 network making this model an excellent choice for industrial applications. Two-stage architectures of different types and kinds achieved moderate performance, while transformer based architectures gave the worst results. EfficientDet-D2 turned out to be the smallest model in terms of the number of parameters.

Our experiments yet again confirm the great ability of DL-based approaches to detect microbial colonies grown in Petri dishes from RGB images. The biggest challenge here is the need to collect large amounts of balanced data. To train detectors in a fully-supervised manner, data must be properly labelled. However, identification of abnormal colonies grown in a Petri dish can be difficult even for a trained specialist. Additionally, variable lighting conditions can make detection even more difficult, which can be observed in our case for EfficientDet-D2 prediction for unrepresented \textit{bright} samples.

\section{Acknowledgements}
Project “Development of a new method for detection and identifying bacterial colonies using artificial neural networks and machine learning algorithms” is co-financed from European Union funds under the European Regional Development Funds as part of the Smart Growth Operational Program. Project implemented as part of the National Centre for Research and Development: Fast Track (grant no. POIR.01.01.01-00-0040/18).

%
%

\bibliographystyle{splncs04}
\bibliography{samplepaper}

\end{document}